\documentclass[runningheads,a4paper]{llncs}
\usepackage{algorithm}
\usepackage{amsmath}  
\usepackage[noend]{algpseudocode}
\usepackage{amssymb}
\usepackage{graphicx}
\usepackage{hyperref}
\usepackage[colorinlistoftodos]{todonotes}
\usepackage{url}
\usepackage{subfig}
\usepackage{multirow}
\usepackage{longtable}
\usepackage{pifont}
\usepackage{booktabs}

\urldef{\mailsa}\path|{email1, email2 , email3, email4,|
\urldef{\mailsb}\path|email5}@springer.com|    

\renewcommand{\vec}[1]{\mathbf{#1}}

\begin{document}
\title{Automatic Detection of Bowel Disease with Residual Networks
% \thanks{Crohn's MRI dataset provided by Uday Patel and Phillip Lung St. Marks hospital (NW-Trust)}
}
%\author{***}
%\institute{***}
 \author{Robert Holland\inst{1} 
\and Uday Patel\inst{2}
\and Phillip Lung\inst{2}
\and Elisa Chotzoglou\inst{1} \and  \\
Bernhard Kainz\inst{1}
}
 \institute{Imperial College London, Department of Computing, BioMedIA, UK \email{robert.holland15|e.chotzoglou16|b.kainz@imperial.ac.uk} \and
 St Mark' Radiology, London North West University Healthcare NHS Trust, UK
 \email{udaypatel2|philliplung@nhs.net}}

\maketitle
\begin{abstract}
Crohn's disease, one of two inflammatory bowel diseases (IBD), affects 200,000 people in the UK alone, or roughly one in every 500. We explore the feasibility of deep learning algorithms for identification of terminal ileal Crohn's disease in Magnetic Resonance Enterography images on a small dataset. We show that they provide comparable performance to the current clinical standard, the MaRIA score, while requiring only a fraction of the preparation and inference time. Moreover, bowels are subject to high variation between individuals due to the complex and free-moving anatomy. Thus we also explore the effect of difficulty of the classification at hand on performance. Finally, we employ soft attention mechanisms to amplify salient local features and add interpretability.

%\keywords{Crohn's disease  \and Deep learning}
\end{abstract}

\section{Introduction}
\subsection{Motivation}
Most people suffering from Crohn's disease are younger than 35 and the cost of their treatment exceeds \pounds 500 million in the UK alone. Symptoms include inflammation of tissue anywhere along the gastrointestinal tract. However, it is most commonly found in the terminal ileum (where the small and large intestine meet). While there is no cure, early detection can vastly improve quality of life.

A successful algorithm would assist radiologists in more accurate diagnosis and follow-up of Crohn's disease. This would be of particular benefit to radiologists with limited experience of Crohn's disease imaging or who encounter patients with Crohn's disease uncommonly. Such an algorithm could also be used to triage patients so that severe cases can be reviewed more immediately, or to perform a secondary review to the radiologist and flag potentially missed cases.

\subsection{Study Outline and Contributions}
Performance of classification tasks on the bowels is degraded by the intrinsic complexity and noise of the anatomy. While Crohn's disease can inflame the entire gastrointestinal (GI) tract, radiologists typically study the terminal ileum when making a diagnosis \cite{Chang2015IntestinalDisease}.
The first question we consider is the extent to which is it possible to classify IBD Crohn's disease from an MRI volume using vanilla deep learning methods. To establish this baseline, we first localise to the ROI using the patient-specific coordinates of the terminal ileum provided by a radiologist. We demonstrate that this semi-automatic technique peforms comparably to the current standard for evaluating Crohn's with MRI, the MaRIA score \cite{Rimola2009MagneticDisease}, while requiring only a fraction of the preprocessing. We also explore how the difficulty and inflammation severity of a sample affects classification performance.

The assumption will then be dropped, such that we are forced to work only with population-specific knowledge, resulting in  weaker localisation. Precision and recall degrade as the now fully-automatic algorithm encounters a worse signal-to-noise ratio (SNR). Finally, we show that in the absence of overfitting soft attention mechanisms \cite{Schlemper2018Attention-GatedDetection} improve performance through amplification of salient local features.

\section{Related Work}
\label{relatedwork}
Currently there are no deep learning methods deployed in the clinic to assist diagnosis of Crohn's disease. Diagnosis is determined entirely by radiologists and clinical professionals who employ various \textit{in vivo} and imaging techniques. Thus, our classification performance will be compared with the clinical standard, the MaRIA score. 
For their similarity in physical domain, we then review similar applications of deep learning to the abdomen.

\subsection{Clinical standards for evaluating IBD}
\label{relatedword:clinical}
The first methods to standardise diagnosis of IBDs were endoscopic scoring systems, such as the Crohn's Disease Endoscopic Index of Severity (CDEIS). However, these incur practical issues; regular endoscopic examinations have several drawbacks related to `\textit{invasiveness, procedure-related discomfort, risk of bowel perforation and relatively poor patient acceptance}' \cite{Rimola2009MagneticDisease}. In fact, a meta-analysis of prospective studies has shown both MRE and CT to have a sensitivity and specificity of greater than 90\% in diagnosing IBDs. To evaluate the MRI, radiologists visually examine the bowels slice by slice and look for high level features. Signs indicative of IBD include increase in T2 signal and thickness of the bowel walls. Rimola et al. \cite{Rimola2009MagneticDisease} developed a scoring system, the MaRIA score, by first extracting these standardised imaging features through manual annotation by a radiologist and then fitting them in a regression model. MaRIA score was found to have a strong correlation with CDEIS. For the detection of disease activity it scored \textbf{0.81} for sensitivity and \textbf{0.89} for specificity.

Challenges in computing the MaRIA score include differentiation of diseased segments from those that are collapsed, variability of disease presentation and image degradation caused by motion \cite{DonaghPotentialObjectives}. Additionally, the aforementioned metrics used in the MaRIA score must be calculated by a radiologist in the terminal ileum, the transverse, ascending, descending and sigmoid colon and the rectum, which is a timely and costly procedure.

\subsection{Machine learning for the automated detection of IBD}
Machine learning can automatically extract local features in the presence of noise, and combine them to make more complex decisions. Thus, it promises to automate the collection of low level features and, as we determine in this work, the diagnosis. Some attempt has been made to automate the collection of features specifically for calculation of the MaRIA score; in 2013 Peter Schüffler et al. \cite{Mahapatra2013AutomaticMRI} used random forests to segment diseased bowels. However, this technique first requires a radiologist to indicate the section of diseased bowel to evaluate. Moreover at the time of the study it required one hour per patient. As far as we can see, there are no studies that use deep learning to directly diagnose IBD from imaging data. Moreover, there are comparatively few medical imaging challenges that focus on the abdomen (notably KiTS19 and CHAOS19) compared to other domains, and as far as we can see, none that regard IBD.

Typically, the medical imaging community has been more focused on tasks such as tumour, lesion and anatomical segmentation. This is evidenced in `A Survey of Deep Learning in Medical Image Analysis' \cite{Litjens2017}, detailing that `\textit{Most papers on the abdomen aimed to localize and segment organs, mainly the liver, kidneys, bladder, and pancreas}'. A more recent review paper, `An overview of deep learning in medical imaging focusing on MRI' \cite{Lundervold2019AnMRI}, describes continued progress in segmentation, registration and image synthesis, but regarding diagnosis and prediction it advises to consult the list from the previous review \cite{Litjens2017} indicating that the main focus still lies in segmentation. Indeed, newer studies on the task of prediction and diagnosis concentrate on the brain, kidney, prostate and spine, but do so via segmentation rather than direction prediction. 

Thus, it may be the case that the optimal method for diagnosing Crohn's IBD operates by first segmenting the terminal ileum. Abdominal segmentation has been attempted, though not including the terminal ileum \cite{Fu2018ARadiotherapy}; dice scores were high for larger anatomy (e.g. liver at $95.3 \pm 0.7$) but significantly reduced for smaller anatomy similar in function to the terminal ileum (e.g. duodenum at $65.5 \pm 8.9$). Furthermore, they go on to describe the limitations of CNNs for inference in the bowels, commenting that `\textit{It is very challenging for the CNN to learn stable representative features for the digestive organs because the appearances, shapes, and sizes of these organs are highly unstable from day to day depending on different food intake and digestion process}' \cite{Fu2018ARadiotherapy}. 

To summarise, there are no studies making direct diagnosis of IBD using deep learning on images. Furthermore, there are also no learning algorithms since the random forests \cite{Mahapatra2013AutomaticMRI} diagnosing IBD from MR volumes. Segmentation is typically preferred to direct diagnosis due to the increased dimensionality of the annotations. As such, we compare our baseline performance to the reported binary classification performance of the MaRIA score in classification of Crohn's disease.

\section{Data}
MRI data has been acquired on a Philips Achieva 1.5 T MR System with acquisition parameters as outlined in 
Table~\ref{tab:MRI}. Use of de-identified data has been consented by the local ethics committees. 

\begin{table}[]
\begin{tabular}{l|p{3.2cm}cccccr}
Planes  & Sequence                                    & FOV [mm] & TR/ TE  & Slice [mm] & Matrix  & NSA & Time [s]   \\
\midrule
Axial   & e-THRIVE (T1 FFE / TFE)                       & 375    & 5.9/3.4 & 3     & 212x160 & 1                          & 20.7 $\times$ 2 \\
Coronal & Single Shot  TSE (T2 TSE) & 375    & 554/120 & 3      & 300x213 & 1                          & 21.1     \\
Axial   & Single Shot  TSE (T2 TSE)                   & 375    & 587/120 & 3.5    & 304x255 & 1                          & 22.3 $\times$ 2
\end{tabular}
\caption{MRI acquisition parameters. Number of signal averages (NSA); Turbo spin echo (TSE)}
    \label{tab:MRI}
\end{table}

The Crohn's MRI dataset is divided into healthy, mild, moderate and severe (with fistulation) terminal ileal inflammation. These represent severity levels 0, 1, 2 and 3 respectively which were originally calculated using the MaRIA score. As there are no terminal ileal ground-truth segmentations available, the only other annotation is the centroid coordinates of the terminal ileum.

Individuals are ranked by classification \textit{difficulty}; an ordering determined by the radiologists who annotated the data. While we cannot formally describe how an MRI volume of a patient might be \textit{difficult} to annotate, we can theorise that it means the symptoms of Crohn's disease are hard to spot or are borderline. These difficulties may correspond to those discussed in computing the MaRIA score discussed in section \ref{relatedword:clinical}. Indeed, we see that as the severity of inflammation decreases the difficulty increases in Table \ref{tab:crohns} (average difficulty is 35.0).

\vspace{-0.3cm}
\begin{table}
    \centering
    \begin{tabular}{c|c|c}
        Inflammation class & Frequency & Average difficulty\\ \hline
        Healthy & 100 & N/A \\
        Mild  & 34 & 39.1\\
        Moderate & 29 & 35.3\\
        Severe & 7 & 19.1\\
    \end{tabular}
    \caption{Distribution of inflammation and suggested classification difficulty}
    \label{tab:crohns}
\end{table}
\vspace{-0.6cm}

\begin{figure}
\vspace{0.5cm}
    \begin{minipage}[t]{.7\textwidth}
        \vspace{-0.5cm}
        \includegraphics[width=0.6\linewidth]{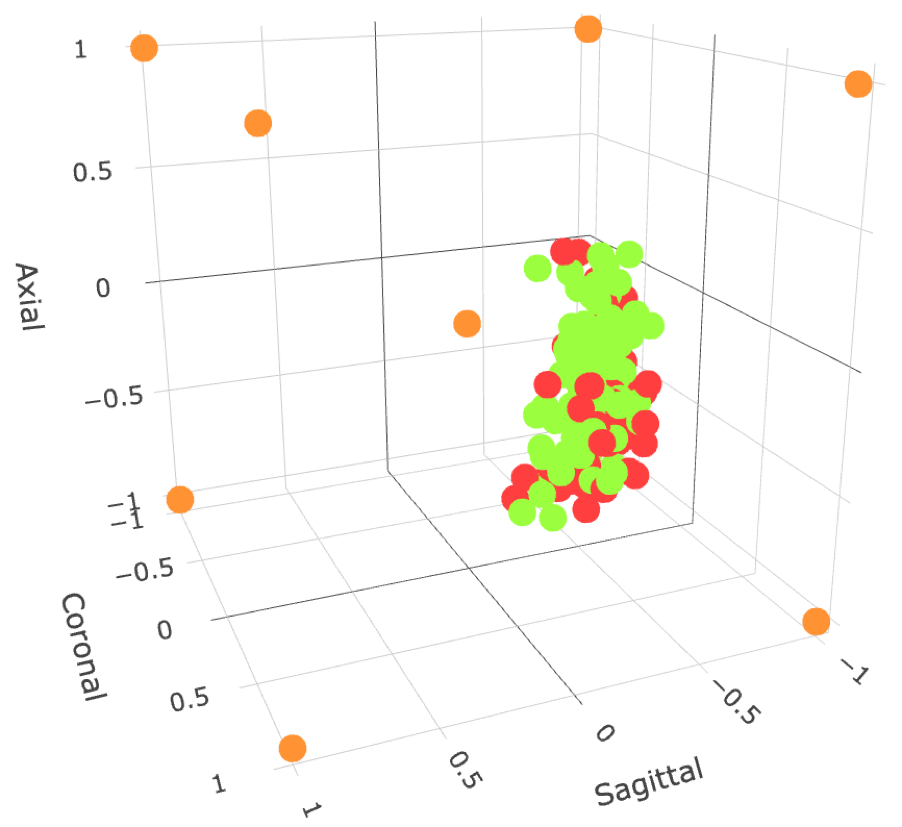}
    \end{minipage}%
    \hspace{-1.0cm}
    \begin{minipage}[t]{0.3\textwidth}
        \vspace{1.1cm}
        \setlength\arraycolsep{1.0pt}
        \small{\small{
         \[\hat{\vec{\mu}} = \begin{bmatrix}
        -0.192\\
        -0.171\\
        -0.111\\
        \end{bmatrix}\]\[
        \hat{\Sigma} =\\ \begin{bmatrix}
        0.012&& -0.005&& -0.014\\
        -0.005&&  0.019 && 0.017\\
        -0.014 && 0.017 && 0.042\\
        \end{bmatrix}\] }}
    \end{minipage}
\caption{Terminal ileal population distribution (normalised to [-1, 1])}
\label{fig:ti_dist}
\end{figure}

Formally, let $\{\vec{i}_j\}_{j=1}^N, \{\vec{d}_j\}_{j=1}^N \text{ such that } \forall j \text{ } \vec{i}_j, \vec{d}_j \in \mathbb{R}^3$ be the set of physical locations of the terminal ilea and the dimensions of the $j^{th}$ patient respectively. Then let the proportional ileal location be $\vec{p}_j = \frac{\vec{i}_j}{\vec{d}_j}$ and suppose that
\[\forall j \text{ } \vec{p}_j \sim \mathcal{N}(\vec{\mu},\,\Sigma)\]
The distribution of $\{\vec{p}_j\}_{j=1}^N$ is shown in Figure \ref{fig:ti_dist}. Given that $\hat{\vec{\mu}} < 0$ and $\hat{\Sigma}$ is small we observe that the terminal ileum is usually confined to one octant of the volume. From this distribution we can define a bounding box that we expect to contain all ilea. We make use of this assumption in preprocessing (see section \ref{application_variants}). We also observe from $\hat{\Sigma}$ that most variation is in the axial direction. This is expected as the method by which we determine the patient's size is most uncertain in this direction (patient dimensions were determined by region growing).

\subsection{Application variants}
\label{application_variants}
We can localise to the ROI using the coordinates of the terminal ileum by extracting a small surrounding window, resulting in the \textit{Localised} dataset. However, in the fully-automatic variant, we are forced to extract a larger region using the estimated distribution shown in Figure \ref{fig:ti_dist} resulting in the \textit{Generic} dataset. The effect of localisation strength on performance is detailed in section \ref{results}. Localisation is crucial for mitigating overfitting, but also for permitting larger batch sizes, since the \textit{Generic} and \textit{Localised}  techniques result in 95.8\% and 99.4\% volume reductions respectively.

% \begin{table}
% \centering
% \begin{tabular}{p{1.5cm}p{2.7cm}p{5cm}p{1.7cm}}
% \textbf{Dataset Alias}& \textbf{Terminal Ileal Coordinates} & \textbf{Description} & \textbf{Volume Reduction}\\ \hline
% Localised Region & \hspace{1.2cm} \ding{51} & Volume of constant size taken about exact ileal coordinates & 99.4\% \\ \hline
% Generic Region & \hspace{1.2cm} \ding{55} & Region growing followed by generic ileal region localisation & 95.8\% \\ \hline
% \end{tabular}
% \caption{The two datasets, localised and generic region}
% \label{preproc:dataset_table}
% \end{table}

\section{Method}
\vspace{-0.1cm}
We are interested in the binary classification power of vanilla deep learning frameworks. As such, due to its efficient use of parameters, we chose ResNet \cite{He} - this affords us larger batch sizes, which are restricted by the dimensionality of the scans. Our custom Resnet uses exclusively $3^3$ filters and ReLU activation. Refer to our network specification in Table \ref{resnetconfig}. Each set of residual blocks, $\vec{d}_j$, begins with a downsampling layer via strided convolution. The residual blocks are followed by a classification module, comprising a global average pooling layer which allows us to feed inputs of variable size to the network. It also reduces the number of learnable parameters in the model as it is followed by a dropout fully connected layer resulting in two output neurons, as in binary classification.

We also add soft attention layers as described in Attention-gated Sononet by Schlemper et al. \cite{Schlemper2018Attention-GatedDetection}. These act as a gate for signal by learning the compatability between pixel-wise features at a large scale and more global, discriminative features taken before the final soft-max layer. This is then normalised to form the attention map (see Figure \ref{fig:attention_maps} for examples) and the dot product is taken with the pixel-wise features to produce attended features. These too pass through a classification module and their prediction is weighted against that of the original network's. To extend our custom Resnet we add an attention layer before the final downsampling layer. This multi-scale technique assists the network in identifying local, salient features such as the terminal ileum and is shown to improve performance in the absence of overfitting.

\begin{table}
\centering
\begin{tabular}{cccc}
\textbf{Layer}  & \textbf{channels}   & \textbf{blocks} & \textbf{resultant feature map} \\ \hline
$\vec{d}_1$  & 64  & 3 & 16x44x44  \\
$\vec{d}_2$  & 128 & 3 & 8x22x22   \\
$\vec{d}_3$  & 256 & 3 & 4x11x11   \\ \hline
global average pooling &     &   & 256       \\\hline
dense layer            &     &   & 2         \\ \hline
\end{tabular}
\caption{Our ResNet configuration for input volume of size 31x87x87}
\label{resnetconfig}
\vspace{-0.6cm}
\end{table}
% \vspace{-1.5cm}

\subsection{Training and Evaluation}
\vspace{-0.1cm}
Loss is computed as cross entropy between the logits and the ground truth labels. We use Adam with $\beta_1 = 0.9, \beta_2 = 0.99$ for the first and second order moment coefficients respectively, and a learning rate of $5 \cdot 10^{-6}$. Due to the reduction in volume (see section \ref{application_variants}) we can use batch sizes of 64, a significant portion of the training set, which somewhat mitigates the intrinsic sample variability by producing more accurate gradient estimates.

Most deep learning frameworks were designed to train on vast datasets. Since we have many millions of parameters but relatively few samples, augmentation is necessary to artificially inflate the dataset. We capture variation present in anatomy and acquisition by including a mix of rotation (about the axial plane), horizontal flipping and random cropping.

All results are determined by four-fold cross validation on stratified training and testing sets, allowing us to evaluate the network on the entire dataset. The limited size of the dataset introduces an upper bound on overall binary classification accuracy of 92.45\% (p-value 0.05), and just 84.8\% on a single fold.
\newpage

\section{Results}
\label{results}
Metrics were recorded when the loss was lowest for each fold. We will refer to Table \ref{cross_fold_results}, containing the results for the combined predictions over the whole dataset, as well as the best performing fold, and detailing the effect of the attention mechanism. It also compares the two levels of localisation that distinguish the \textit{Localised} and \textit{Generic} datasets.

\vspace{-0.1cm}
\begin{table}
\centering
\begin{tabular}{crcccc}
 & & \multicolumn{2}{c}{\textbf{Generic Region}}\hspace{2mm} & \multicolumn{2}{c}{\textbf{Localised Region}} \\ \cline{1-6} 
\textbf{Attention} &  & A  & H   & A& H    \\ \hline
\multirow{2}{*}{\textbf{\ding{55}}} & \textbf{Average}  &  0.61/0.20  & 0.62/0.91 &  0.76/0.69 & 0.79/0.85 \\
& \textbf{Best} &  0.73/0.47 & 0.71/0.88 & 0.93/0.82 & 0.89/0.96 \\ \hline
\multirow{2}{*}{\textbf{\ding{51}}} & \textbf{Average}  &    0.59/0.14  & 0.61/0.93 &  \textbf{0.79}/\textbf{0.80} & \textbf{0.86}/\textbf{0.85} \\
& \textbf{Best} & 0.60/0.35 & 0.66/0.84 & \textbf{0.94}/\textbf{0.94} & \textbf{0.96}/\textbf{0.96}
\vspace{0.15cm}
\end{tabular}
\caption{Best and average cross-fold binary classification performance for all application variants (formatted by precision/recall, and where A and H denote the abnormal and healthy classes respectively)}
\label{cross_fold_results}
\end{table}
\vspace{-0.9cm}

In most cases performance is reduced on the underrepresented class of abnormal patients. Moreover, performance is signficantly increased on localised data, and achieved best performance with attention mechanisms - this variant achieves weighted f-1 score \textbf{0.83}, demonstrating a strong correlation with the MaRIA score.

However, there is a large disparity between the best fold and the cross-fold average. In fact, the performance of any given fold was found to be highly dependent on the difficulty of the test set. Here we consider the difficulty of the abnormal samples only, assuming that healthy individuals present similar difficulty. We find that difficulty of the best fold was merely 31.3 while the worst was 42.3. Moreover, for the \textit{Localised} variant with attention mechanisms, the average difficulty of incorrectly predicted abnormals was high, at 51.78, and of the seven severely inflamed individuals none were incorrectly classified (see Table \ref{tab:class_performance}). Classification power consistently increases with inflammation severity.

The limited size of the dataset introduced severe overfitting in training, forcing us to restrict the depth of the network and degrading overall performance. Furthermore, larger networks performed worse on the \textit{Generic}, or population-specific, variant due to the reduction in SNR. This introduced difficulties in comparing variants on a standardised architecture.

\vspace{-0.4cm}
\begin{table}
    \centering
    \begin{tabular}{c|cccc}
        \textbf{Inflammation} & Severe & Moderate & Mild & Healthy\\ \hline
        \textbf{Accuracy (\%)} & 100.0 & 86.2 & 70.6 & 85.0
    \end{tabular}
    \caption{Classification accuracy of best performing variant per inflammation class (for class support refer to Table \ref{tab:crohns})}
    \label{tab:class_performance}
\end{table}
\vspace{-1.2cm}

\newpage
\subsection{Attention}
\begin{figure}
\centering
\subfloat[Zero-padding] {\includegraphics[width=0.1695\linewidth]{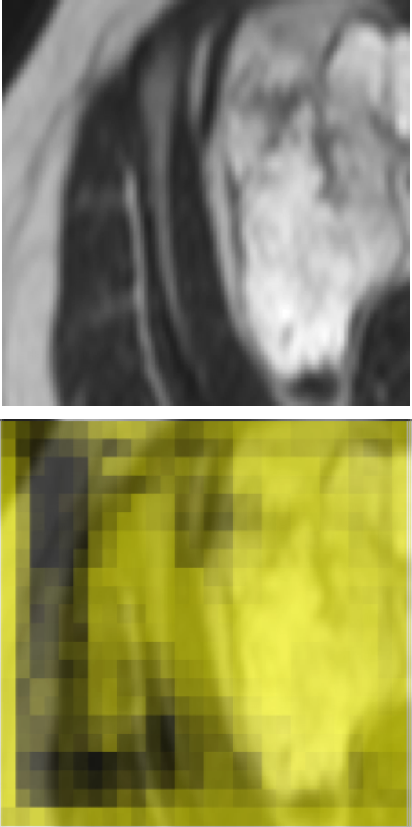}
\label{fig:zeropad}}
\quad
\subfloat[Improved localisation with mirror-padding] {\includegraphics[width=0.71\linewidth]{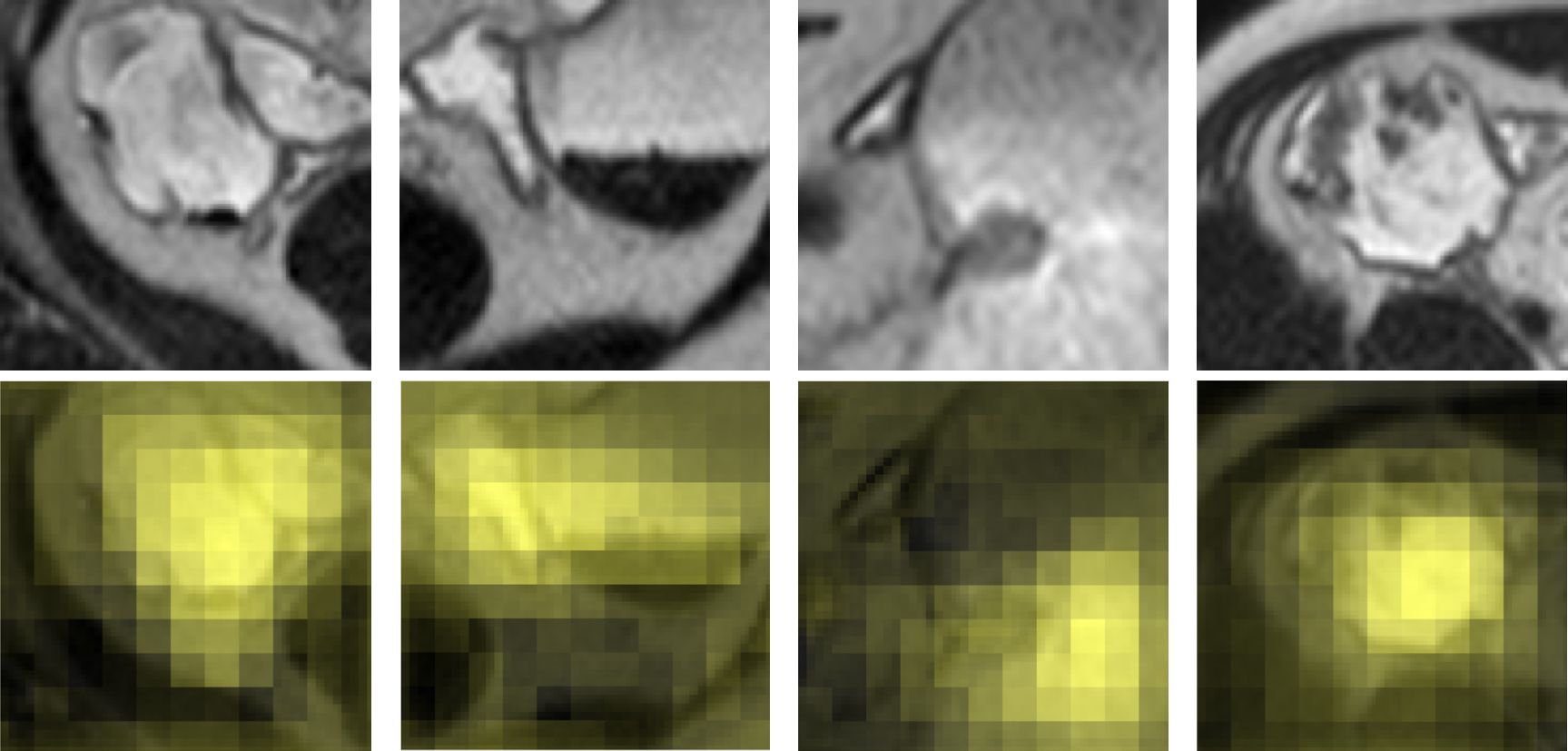}
\label{fig:mirrorpad}}
\quad
\caption{Attention maps on the \textit{Localised} dataset (original slice and with attention overlayed on top and bottom rows respectively)}
\label{fig:attention_maps}
\end{figure}

\vspace{-0.0cm}
Attention mechanisms were found to exacerbate overfitting in scenarios with a low SNR but otherwise boosted performance. This can be seen by observing that attention boosts performance on the \textit{Localised} dataset but degrades it on \textit{Generic}. %——- ADDED MATERIAL
We theorise that attention mechanisms can only become effective techniques to identify salient, local features within a network if the additional parameters they introduce are not accidentially misused for overfitting. There is evidence for this since the lowest cross entropy achieved by the best performing fold on the \textit{Generic} dataset increased from 0.565 to 0.619 with the addition of attention mechanisms. %—— END OF ADDED MATERIAL
Referring to Figure \ref{fig:attention_maps}, it also assisted us in \textit{debugging} our network by highlighting that zero-padding allows the network to localise to regions that can be overfit on, such as bordering tissue (see Figure \ref{fig:zeropad}); mirror padding solves this issue. From Figure \ref{fig:mirrorpad} we deduce that the attention mechanism successfully identifies the relevant bowel section, reinforcing our confidence in the diagnoses.

\vspace{-0.5cm}
\vspace{-0.2cm}

\section{Discussion and Conclusion}
\vspace{-0.2cm}
In this work we demonstrated that a generic deep learning network, trained on a very small MRI dataset, correlates strongly to the MaRIA score, the current clinical standard, while requiring a fraction of the preprocessing by the radiologist. However, the framework is not without limitations in that performance is highly dependent on the level of localisation used in preprocessing and the difficulty rating of the classification at hand. Furthermore, the low dimensionality of the output variable introduces statistical upper bounds on classification power and increases overfit. In this paper the evaluation criteria is solely based on expert radiologist assessment of the MRI data. Validation through colonoscopy  is subject to future work, pending ethical approval.

Despite this, we observed very high classification power on the moderate to severely inflamed individuals, suggesting that this algorithm could provide secondary diagnoses to the radiologist in order to flag potentially missed cases. Overall, this pilot study highlights that deep learning is a very promising technique as a method for diagnosing disease in the bowels, and indicates that a larger dataset should continue to be collected for further evaluation. Finally, the limitations encountered through predicting low-dimensionality data might be alleviated by instead automating segmentation of the terminal ileum using deep learning, as a precursor to diagnosis. Thus, we recommend that terminal ileal ground-truth segmentations also be collected.

\noindent\textbf{Acknowledgements:} This research was kindly supported by Intel and hardware donations from Nvidia.

\bibliographystyle{splncs04}
\bibliography{main}

\begin{thebibliography}{1}
\providecommand{\url}[1]{\texttt{#1}}
\providecommand{\urlprefix}{URL }
\providecommand{\doi}[1]{https://doi.org/#1}

\bibitem{Chang2015IntestinalDisease}
Chang, C.W., Wong, J.M., Tung, C.C., Shih, I.L., Wang, H.Y., Wei, S.C.:
  {Intestinal Stricture in Crohn's Disease}. Intestinal Research
  \textbf{13}(1), ~19 (1 2015). \doi{10.5217/ir.2015.13.1.19}

\bibitem{DonaghPotentialObjectives}
Donagh, C., Walshe, T.M., Roche, C., Lohan, D., Cronin, C.G., Murphy, J.:
  {Potential Pitfalls in MRI enterography-A pictorial review Learning
  objectives} . \doi{10.1594/ecr2012/C-2046}, \url{www.myESR.org}

\bibitem{Fu2018ARadiotherapy}
Fu, Y., Mazur, T.R., Wu, X., Liu, S., Chang, X., Lu, Y., Li, H.H., Kim, H.,
  Roach, M.C., Henke, L., Yang, D.: {A novel MRI segmentation method using
  CNN-based correction network for MRI-guided adaptive radiotherapy}. Medical
  Physics  \textbf{45}(11),  5129--5137 (11 2018). \doi{10.1002/mp.13221}

\bibitem{He}
He, K., Zhang, X., Ren, S., Sun, J.: {Deep residual learning for image
  recognition}. In: Proceedings of the IEEE Conference on Computer Vision and
  Pattern Recognition. pp. 770--778 (2016)

\bibitem{Litjens2017}
Litjens, G., Kooi, T., Bejnordi, B.E., Arindra, A., Setio, A., Ciompi, F.,
  Ghafoorian, M., Laak, J.A.W.M.V.D., Ginneken, B.V., S{\'{a}}nchez, C.I.: {A
  survey on deep learning in medical image analysis}  \textbf{42}(December
  2012),  60--88 (2017). \doi{10.1016/j.media.2017.07.005}

\bibitem{Lundervold2019AnMRI}
Lundervold, A.S., Lundervold, A.: {An overview of deep learning in medical
  imaging focusing on MRI}. Zeitschrift f{\"u}r Medizinische Physik
  \textbf{29}(2),  102--127 (2019)

\bibitem{Mahapatra2013AutomaticMRI}
Mahapatra, D., Schuffler, P.J., Tielbeek, J.A., Makanyanga, J.C., Stoker, J.,
  Taylor, S.A., Vos, F.M., Buhmann, J.M.: {Automatic detection and segmentation
  of Crohn's disease tissues from abdominal MRI}. IEEE Transactions on Medical
  Imaging  \textbf{32}(12),  2332--2347 (12 2013).
  \doi{10.1109/TMI.2013.2282124}

\bibitem{Rimola2009MagneticDisease}
Rimola, J., Rodriguez, S., Garc{\'{i}}a-Bosch, O., Ord{\'{a}}s, I., Ayala, E.,
  Aceituno, M., Pellis{\'{e}}, M., Ayuso, C., Ricart, E., Donoso, L.,
  Pan{\'{e}}s, J.: {Magnetic resonance for assessment of disease activity and
  severity in ileocolonic Crohn's disease}. Gut  \textbf{58}(8),  1113--1120 (8
  2009). \doi{10.1136/gut.2008.167957}

\bibitem{Schlemper2018Attention-GatedDetection}
Schlemper, J., Oktay, O., Schaap, M., Heinrich, M., Kainz, B., Glocker, B.,
  Rueckert, D.: Attention gated networks: Learning to leverage salient regions
  in medical images. Medical Image Analysis  \textbf{53},  197 -- 207 (2019).
  \doi{https://doi.org/10.1016/j.media.2019.01.012},
  \url{http://www.sciencedirect.com/science/article/pii/S1361841518306133}

\end{thebibliography}

% \printbibliography

\end{document}